\documentclass[sigconf]{acmart} %

\AtBeginDocument{
  \providecommand\BibTeX{{
    \normalfont B\kern-0.5em{\scshape i\kern-0.25em b}\kern-0.8em\TeX}}}

\setcopyright{acmcopyright}
\copyrightyear{2018}
\acmYear{2018}
\acmDOI{10.1145/1122445.1122456}

\acmConference[Chengdu '21]{Chengdu '21: ACM MULTIMEDIA}{October 20--24, 2021}{Chengdu, China}
\acmBooktitle{Chengdu '21: ACM Multimedia,
  October 20--24, 2021, Chengdu, China}
\acmPrice{15.00}
\acmISBN{978-1-4503-XXXX-X/18/06}

\usepackage{soul}

\usepackage{hyperref}
\usepackage[utf8]{inputenc}
\usepackage{caption}
\usepackage{graphicx}
\usepackage{amsmath}
\usepackage{booktabs}
\usepackage{multirow}
\usepackage{hyperref}
\usepackage{latexsym} 
\usepackage[ruled,linesnumbered]{algorithm2e}
\usepackage{makecell}
\DeclareMathAlphabet{\mathcal}{OMS}{cmsy}{m}{n}

\usepackage{amsmath,amsfonts,bm}

\def\eqref#1{equation~\ref{#1}}

\def\1{\bm{1}}

\def\va{{\bm{a}}}

\def\vd{{\bm{d}}}
\def\ve{{\bm{e}}}

\def\vm{{\bm{m}}}

\def\vs{{\bm{s}}}
\def\vt{{\bm{t}}}
\def\vu{{\bm{u}}}
\def\vv{{\bm{v}}}

\def\vx{{\bm{x}}}
\def\vy{{\bm{y}}}
\def\vz{{\bm{z}}}

\DeclareMathAlphabet{\mathsfit}{\encodingdefault}{\sfdefault}{m}{sl}
\SetMathAlphabet{\mathsfit}{bold}{\encodingdefault}{\sfdefault}{bx}{n}

\usepackage{color, colortbl}
\usepackage{import}
\usepackage{balance}
\begin{document}

\title{Mitigating Generation Shifts for Generalized Zero-Shot Learning
}
\author{Zhi Chen$^{1}$, Yadan Luo$^{1}$, Sen Wang$^{1}$, Ruihong Qiu$^{1}$, Jingjing Li$^{2}$, Zi Huang$^{1}$}
\affiliation{
  \institution{$^1$The University of Queensland, $^{2}$University of Electronic Science and Technology of China}
}
\email{ {zhi.chen, y.luo, sen.wang, r.qiu}@uq.edu.au, lijin117@yeah.net, huang@itee.uq.edu.au}

\begin{abstract}
Generalized Zero-Shot Learning (GZSL) is the task of leveraging semantic information (e.g., attributes) to recognize the seen and unseen samples, where unseen classes are not observable during training. It is natural to derive generative models and hallucinate training samples for unseen classes based on the knowledge learned from the seen samples. However, most of these models suffer from the generation shifts, where the synthesized samples may drift from the real distribution of unseen data. In this paper, we propose a novel Generation Shifts Mitigating Flow framework, which is comprised of multiple conditional affine coupling layers for learning unseen data synthesis efficiently and effectively. In particular, we identify three potential problems that trigger the generation shifts, \textit{i.e.}, \textit{semantic inconsistency}, \textit{variance decay}, and \textit{structural permutation} and address them respectively. First, to reinforce the correlations between the generated samples and the respective attributes, we explicitly embed the semantic information into the transformations in each of the coupling layers. Second, to recover the intrinsic variance of the synthesized unseen features, we introduce a visual perturbation strategy to diversify the intra-class variance of generated data and hereby help adjust the decision boundary of the classifier. Third, to avoid structural permutation in the semantic space, we propose a relative positioning strategy to manipulate the attribute embeddings, guiding which to fully preserve the inter-class geometric structure. Experimental results demonstrate that GSMFlow achieves state-of-the-art recognition performance in both conventional and generalized zero-shot settings. Our code is available at: \color{pink}{\href{https://github.com/uqzhichen/GSMFlow}{https://github.com/uqzhichen/GSMFlow}}

\end{abstract}

\ccsdesc[500]{Computing methodologies ~ Computer vision}

\keywords{Generalized zero-shot learning, conditional generative flows}

\maketitle

\section{Introduction}
Deep learning techniques have significantly boosted the performance of many tasks in computer vision \cite{luo2018robust,luo2018coarse,wang2020prototype,luo2020learning,luo2020adversarial,wang2020human,luo2019curiosity,luo2020collaborative,zhang2021privacy,zhang2021aggregation,zhang2021high,zhang2017semi}. However, the performance gains come at the cost of an enormous amount of labeled data. Generally, shallow classification algorithms need a certain amount of training data of every class, while deep learning dramatically amplifies the need. Data labeling is time-consuming and expensive, even if the raw data is usually plentiful. Moreover, it is unrealistic to require labeled data for every class, and therefore much attention from researchers has been drawn to Zero-Shot Learning (ZSL) as a solution \cite{akata2015label,xian2016latent,akata2015evaluation,yang2016zero,li2019zero}. By incorporating side information, \textit{e.g.,} class-level semantic attributes, ZSL transfers semantic-visual relationships from the seen classes to unseen classes without any visual samples. While conventional ZSL aims to recognize only unseen classes, it is infeasible to assume that we will only come across samples from unseen classes. Hence, in this paper, we consider a more realistic and challenging task, Generalized Zero-Shot Learning (GZSL), to classify over both seen and unseen classes \cite{huang2019generative,schonfeld2019generalized,xian2018feature,li2020learning}.

\begin{figure}[t]
\centering
\includegraphics[width=85mm]{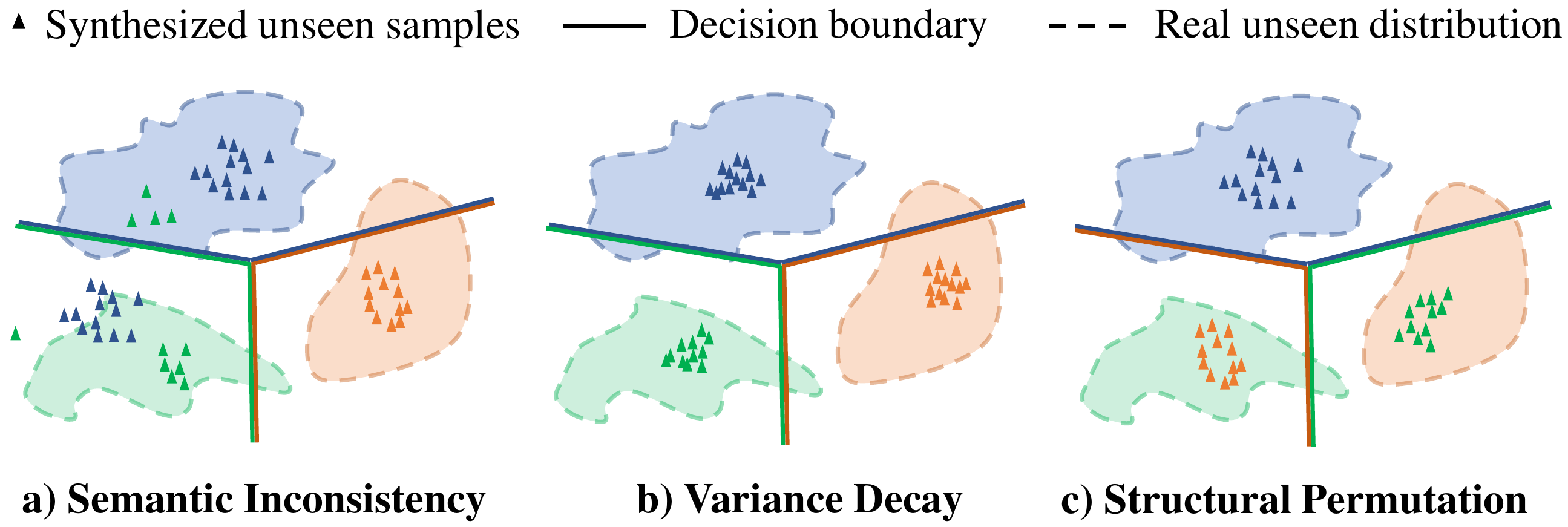}
\caption{ An illustration of the generation shifts. 
a) Implicit semantic encoding causes the semantic inconsistency when generating unseen samples. b) The synthesized samples collapse to fixed modes. c) The synthesized samples fail to fully preserve the geometric relationships in semantic space.
}
\label{intro}
\vspace{-10pt}
\end{figure}

GZSL can be roughly categorized into embedding-based methods \cite{xian2016latent,akata2015label,akata2015evaluation} and generative methods \cite{li2019leveraging,xian2018feature,schonfeld2019generalized,li2019alleviating,}. 
Embedding-based methods usually cast the semantic and visual features into the same space, and thus the compatibility scores between the visual features and all classes are computed to make predictions. In contrast, generative methods \cite{long2017zero,xian2018feature,schonfeld2019generalized,chen2020rethinking,shen2020invertible,li2021investigating} cast the GZSL problem into a supervised classification task by generating synthesized visual features for unseen classes. 
Then a supervised classifier can be trained on both real seen visual features and the synthesized unseen visual features. 

\begin{figure*}[t]
\centering
\includegraphics[width=165mm]{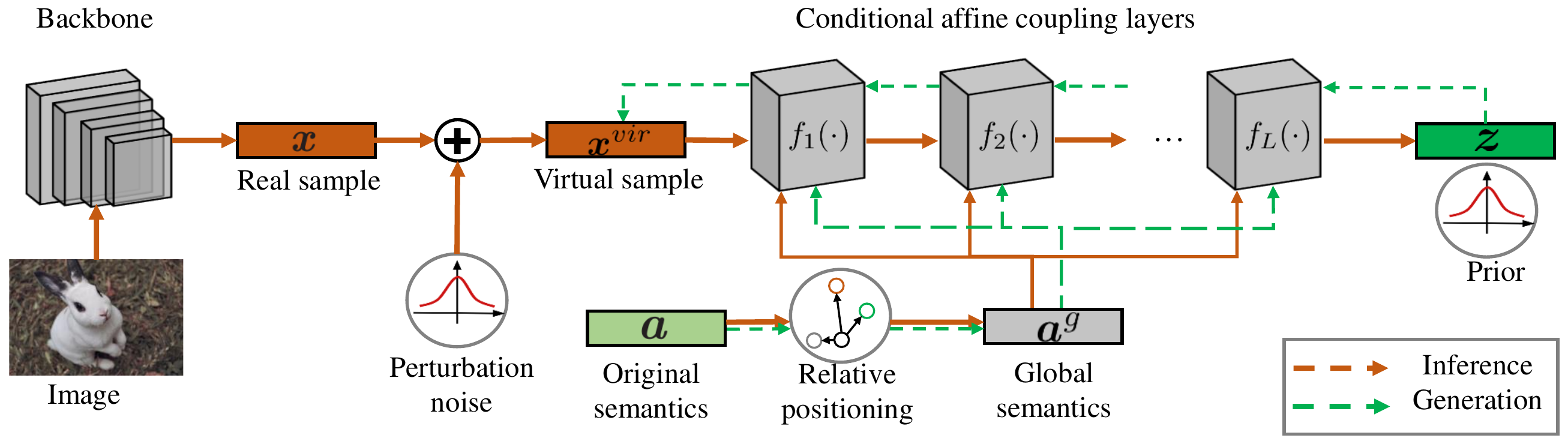}
\caption{An illustration of the proposed GSMFlow framework. The conditional generative flow is comprised of a series of conditional affine coupling layers. Particularly, the perturbation is injected into the original visual features to complement the potential patterns and the global semantics are computed with relative positioning to semantic anchors. For inference, a latent variable $\vz$ is inferred from the visual features of an image sample $\vx$ conditioned on a global semantic vector $\va^g$. Inversely, given $\vz$ drawn from a prior distribution and a global semantic vector $\va^g$, GSMFlow can generate a visual sample accordingly.}
\label{architecture}
\end{figure*}


In spite of the advances achieved by the generative paradigm, the synthesized visual features may not be guaranteed to cover the real distribution of unseen classes, since the unseen features are not exposed during training.
Due to the gap between the synthesized and the real distributions of unseen features, the trained classifier tends to misclassify the unseen samples.
Therefore, it is critical to analyze the causes and mitigate the resulted generation shifts. In this paper, as illustrated in Figure \ref{intro}, we identify three common types of generation shifts in generative zero-shot learning methods:
\begin{itemize}
    \item \textbf{Semantic inconsistency:} The state-of-the-art GZSL approach \cite{shen2020invertible} focuses on preserving the exact generation cycle consistency by a normalizing flow. It constructs a complex prior and disentangle the outputs into semantic and non-semantic vectors. Such implicit encoding may cause the generated samples to be incoherent with the given attributes and deviated from the real distributions.
    \item \textbf{Variance decay:} The synthesized samples commonly collapse into fixed modes and fail to capture the intra-class variance of unseen samples, which can be originated from different poses, illumination and background.
     \item \textbf{Structural permutation:} It is also hard to fully preserve the geometric relationships between different attribute categories in the shared subspace. If the relative position of unseen samples changes in the visual space, the recognition performance will inevitably degenerate. 
\end{itemize}

To address the aforementioned generation shifts, in this paper, we propose a novel framework for generalized zero-shot learning, namely Generation Shifts Mitigating Flow (GSMFlow), as depicted in Figure \ref{architecture}. In particular, to tackle the semantic inconsistency, we explicitly embed the semantic information into the transformations in each of the coupling layers during inference and generation, enforcing the unseen visual feature generation to be semantically consistent. Furthermore, to mitigate the variance decay issue, a visual perturbation strategy is proposed to diversify training samples by dynamically injecting the perturbation noise into the input samples. With this strategy, the decision boundary could be adjusted for suiting the real unseen samples at the test time. Moreover, to alleviate the semantic structural permutation, the relative positioning mechanism is adopted to correct the attribute representations by preserving the global geometrical relationship to the specific semantic anchors. 
Specifically, the visual features $\vx$ of a real sample are first extracted from a backbone network, \textit{e.g.,} ResNet101. In the inference stage, the real sample is perturbed to be a virtual sample. The virtual sample is then input into a conditional generative flow $f(\cdot)$. The inference process is conditioned on the manipulated global semantic vector to infer the latent factors drawn from a prior distribution. The learned generative flow has its inverse transformation as the generative network. Similarly, a global semantic vector is progressively injected into each of the inverse coupling layers to generate unseen visual features from latent factors. We then train a unified classifier with the real seen features and the generated unseen features, which aims to accurately recognize both seen and unseen classes at the test time. To sum up, the contributions of our work are listed as follows:
\begin{itemize}
    \item 
    We propose a novel GSMFlow framework for GZSL, which explicitly incorporates the class-level semantic information into both the forward and the inverse transformations of the conditional generative flow, which encourages the synthesized samples to be more coherent with the respective semantic information.  
    
    \item We propose a visual perturbation strategy to recover the intrinsic variance in the synthesized unseen samples. By injecting perturbation noise into the seen training sample, we diversify the training samples to enrich the original feature space. As the generative flow is exposed to more diverse virtual samples, the synthesized samples of unseen classes can thus capture more visual potentials.
   
    \item To preserve the geometric relationship between different semantic vectors in the semantic space, we choose different semantic anchors to revise the representation of the attributes.
    \item Comprehensive experiments and in-depth analysis on four GZSL benchmark datasets demonstrate the state-of-the-art performance by the proposed GSMFlow framework in the GZSL tasks.
\end{itemize}

The rest of the paper is organised as follows. We briefly review related work in Section 2. GSMFlow is presented in Section 3, followed
by the experiments in Section 4. Lastly, Section 5 concludes the
paper.

\section{Related Work}
\subsection{Traditional Zero-shot Learning}
Traditional solutions towards zero-shot learning are mostly the embedding-based methods. The pioneering method ALE \cite{akata2015label} proposes to employ embedding functions to measure the compatibility scores between a semantic embedding and a data sample. SJE \cite{akata2015evaluation} extends ALE by using structured SVM \cite{tsochantaridis2005large} and takes advantage of the structured outputs. DeViSE \cite{frome2013devise} constructs a deep visual semantic embedding model to map 4096-dimensional visual features from AlexNet \cite{krizhevsky2012imagenet} to 500 or 1000-dimensional skip-gram semantic embeddings. EZSL \cite{romera2015embarrassingly} theoretically gives the risk bound of the generalization error and connects zero-shot learning with the domain adaptation problem. More recently, SAE \cite{kodirov2017semantic} develops a cycle embedding approach with an autoencoder to reconstruct the learned semantic embeddings into a visual space. Later, the cycle architecture is also investigated in generative methods \cite{chen2020canzsl}. However, these early methods have not achieved satisfactory results on zero-shot learning. Particularly, when applying on the GZSL task, the unseen class performance is even worse. Recently, thanks to the advances of generative models, by generating missing visual samples of unseen classes, zero-shot learning can be converted into a supervised classification task.

\subsection{Generative GZSL}
A number of generative methods have been applied for GZSL, \textit{e.g.,} Generative Adversarial Nets (GANs) \cite{goodfellow2014generative}, Variational Autoencoders (VAEs) \cite{kingma2013auto}, and Alternating Back-Propagation algorithms (ABPs) \cite{han2017alternating}. f-CLSWGAN \cite{xian2018feature} presents a WGAN-based \cite{arjovsky2017wasserstein} approach to synthesize unseen visual features based on semantic information. CADA-VAE \cite{schonfeld2019generalized} proposes to stack two VAEs, each for one modality, and aligns the latent spaces. The latent space, thus, can enable information sharing among different modality sources. LisGAN \cite{li2019leveraging} is inspired by the multi-view property of images and improves f-CLSWGAN by encouraging the generated samples to approximate at least one visually representative view of samples. CANZSL \cite{chen2020canzsl} considers the cycle-consistency principle of image generation and proposes a cycle architecture by translating synthesized visual features into semantic information. ABP-ZSL \cite{zhu2019learning} adopts the rarely studied generative models ABPs to generate visual features for unseen classes. GDAN \cite{huang2019generative} incorporates a flexible metric in the model’s discriminator to measure the similarity of features from different modalities. 

The bidirectional conditional generative models enforce cycle-consistent generation and allow the generated images to truthfully reflect the conditional information \cite{zhu2017unpaired,chen2019cycle,chen2020canzsl}. Instead of encouraging cycle consistency through adding additional reverse networks, generative flows are bidirectional and cycle-consistent in nature. The generative flows are designed to infer and generate within the same network. Also, the generative flows are lightweight comparing to other methods as no auxiliary networks are needed, \textit{e.g.,} discriminator for GANs, variational encoder for VAEs. Some conditional generative flows \cite{ardizzone2018analyzing,ardizzone2019guided} are proposed to learn image generation from class-level semantic information. IZF \cite{shen2020invertible} adopts the invertible flow model \cite{ardizzone2018analyzing} for GZSL. It implicitly learns the conditional generation by inferring the semantic vectors. Such implicit encoding may cause the generated samples to be incoherent with the given attributes. Instead, we explicitly blend the semantic information into each of the coupling layers of the generative flow, learning the semantically consistent visual features. We also argue that the MMD regularization in IZF is inappropriate for GZSL, since the seen and unseen classes are coalesced. More explanation is depicted in Section 3.2.


\subsection{Visual Perturbation}
Our proposed visual perturbation approach is similar to data augmentation techniques. In many deep learning tasks, data augmentation \cite{shorten2019survey} is an effective strategy to improve model performance. Generally, bigger datasets better facilitate the learning of deep models \cite{sun2017revisiting}. Common data augmentation techniques on image data include flipping, rotating, cropping, color jittering, edge enhancement, random erasing \cite{mikolajczyk2018data}, etc. Instead of augmenting the raw data samples, DeVries and Taylor \cite{devries2017dataset} propose an augmentation strategy in the features space. In the dense feature space, the augmentation is more direct and unlimited augmented feature samples can be yielded.

Generative zero-shot learning is intrinsically a data augmentation approach to handle the zero-shot learning problem by augmenting the training data of the unseen classes. However, the augmentation power is limited by the training data of the seen classes, as the empirical distribution of training data cannot faithfully reflect the underlying real distribution of seen classes. In this case, we resort to a data augmentation strategy to complement the potential patterns depicted seen classes and further transfer to unseen classes. 

It is noteworthy that the motivation of our approach is also different from these data augmentation techniques. These methods statically augment from the real samples for training, resulting in limited augmented samples. Instead of static augmentation, our perturbation noises are sampled dynamically, leading to unlimited augmented samples.

\section{Methodology}
This section begins with formulating the GZSL problem and introducing the notations. The proposed GSMFlow is outlined next, followed by the visual perturbation strategy and the relative positioning approach to mitigate the generation shifts problem. The model training and zero-shot recognition are introduced lastly.



\subsection{Preliminaries}
Consider two datasets - a seen set $\mathcal{S}$ and an unseen set $\mathcal{U}$. The seen set $\mathcal{S}$ contains $N^s$ training samples $\vx^s \in \mathcal{X}$ and the corresponding class labels $\vy^s$, \textit{i.e.}, $\mathcal{S} = \{ \vx_{(i)}^{s}, \vy_{(i)}^{s} \}_{i=1}^{N^s}$. Similarly, $\mathcal{U} = \{ \vx_{(i)}^{u}, \vy_{(i)}^{u} \}_{i=1}^{N^u}$, where $N^u$ is the number of unseen samples. There are $C_s$ seen and $C_u$ unseen classes, so that $\vy^s \in \{1,\ldots,C_s \}$ and $\vy^u \in \{C_s+1,\ldots,C_u\}$. Note that the seen and unseen classes are mutually exclusive. There are $C_s+C_u$ class-level semantic vectors $\va \in \mathcal{A}$, where the class-level semantic vectors for the seen and unseen classes are $\{\va_{i}^{s}\}_{i=1}^{C_s}$ and $\{\va_{i}^{u}\}_{i=1}^{C_u}$, respectively.
$\va_{i}$ represents the semantic vector for the $i$-th class. In the setting of GZSL, we only have access to the seen samples $\mathcal{S}$ and semantic vectors $\{\va_{i}^{s}\}_{i=1}^{C_s}$ during training. Hence, for brevity, in the demonstration of the training process, we will omit the superscript $s$ for all the seen samples, \textit{i.e.}, $\vx=\vx^s$, $\va = \va^s$.

\subsection{Conditional Generative Flow}
To model the distribution of unseen visual features, we resort to conditional generative flows \cite{ardizzone2019guided}, a stream of simple yet powerful generative models. 

A conditional generative flow learns inference and generation within the same network. Let $\vz \in \mathcal{Z}$ be a random variable from a particular prior distribution $p_{Z}$, \textit{e.g.,} Gaussian, which has the same dimension as the visual features. We denote the inference process as the forward transformation $f(\cdot): \mathcal{X} \times \mathcal{A} \rightarrow \mathcal{Z}$, whose inverse transformation $g(\cdot) = f^{-1}(\cdot) : \mathcal{Z} \times \mathcal{A} \rightarrow \mathcal{X}$ is the generation process. The transformation is composed of $L$ bijective transformations $f(\cdot) = f_1(\cdot) \circ f_2(\cdot) \circ... \circ f_L(\cdot)$, the forward and inverse computations are the composition of the $L$ bijective transformations. 
Then, we can formalize the \textit{conditional generative flow} as:
\begin{equation}
\begin{gathered}
  \vz = f(\vx ; \va, \theta), \quad\vz \sim  \mathcal{N}(\bm{0}, \bm{I}),
\end{gathered}
\end{equation}
where the transformation $f(\cdot)$ parameterized with $\theta$ learns to transform a sample $\vx$ in the target distribution to the prior data distribution $p_Z$ conditioned on the corresponding class-level semantic vector $\va$. The forward transformation $g(\cdot)$ has its inverse transformation $g(\cdot) = f^{-1}(\cdot)$, which flows from the prior distribution $p_Z$ towards the target data distribution $p_X$:
\begin{equation}
\begin{gathered}
  \vx = g(\vz; \va, \theta).
\end{gathered}
\end{equation}
With the \textit{change of variable formula}, the bijective function $f(\cdot)$ can be trained through maximum log-likelihood:
\begin{equation}
\begin{gathered}
  \log p_{X}(\vx;\va, \theta) =\log p_{Z}(f(\vx;\va, \theta)) + \log\Big\lvert \text{det}(\frac{\partial f}{\partial \vx} ) \Big\rvert,
\end{gathered}
\end{equation}
where the latter half term denotes the logarithm of the determinant of the Jacobian matrix.

According to Bayes' theorem, the posterior distribution $p_{X}(\theta ; \vx,\va)$ for the parameter $\theta$ is proportional to $p_{X}(\vx;\va,\theta) p_{\theta}(\theta)$. The objective function can then be formulated as:
\begin{equation}
\begin{gathered}
  \mathcal{L}_{flow} = \mathbb{E}[-\log p_{X}(\vx;\va,\theta)] - \log p_{\theta}(\theta),
  \label{flow}
\end{gathered}
\end{equation}

\begin{figure}[t]
\centering
\includegraphics[width=85mm]{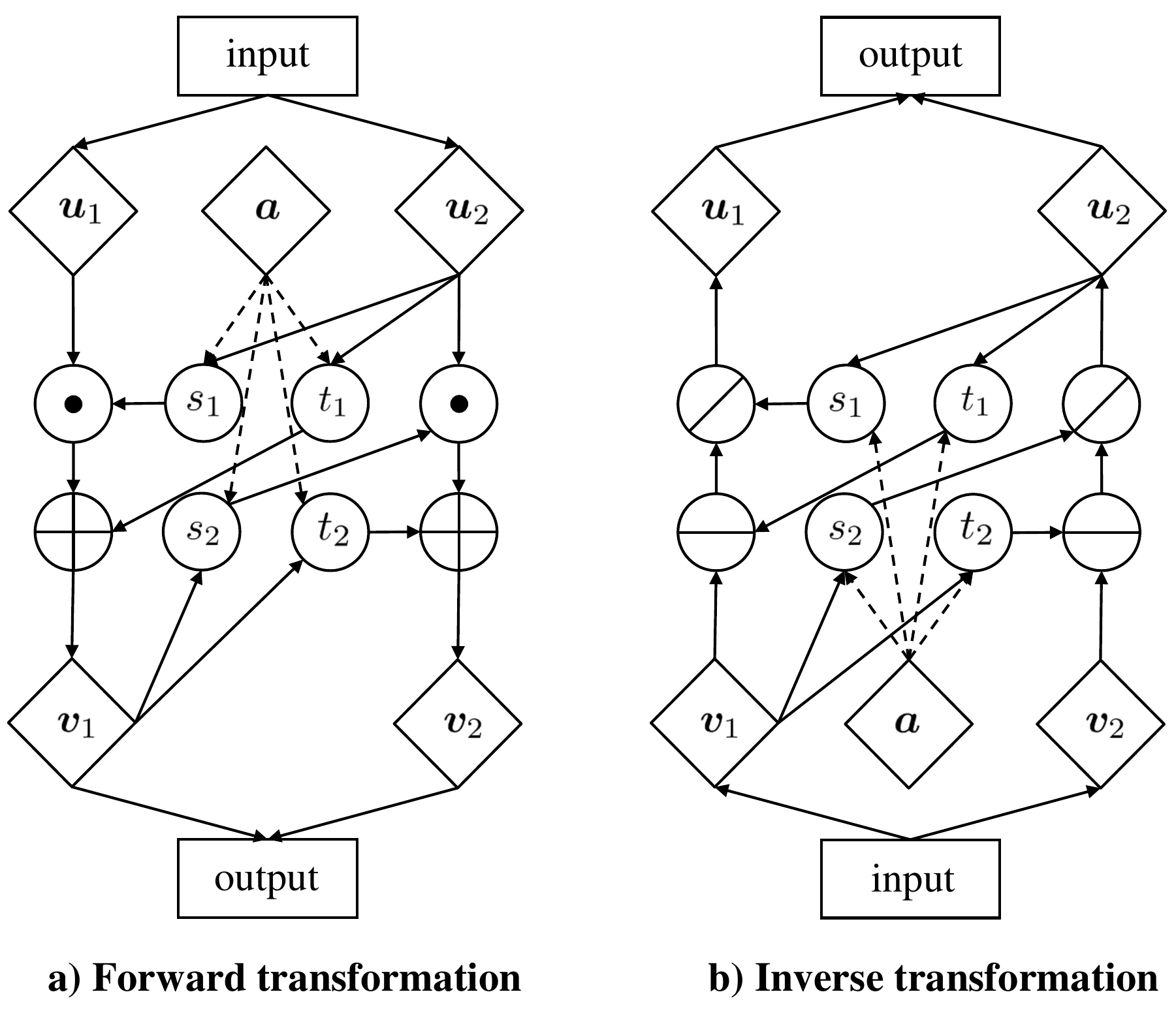}
\caption{The flowcharts of the conditional affine coupling layers. (a) The transformation flowchart when training the model. (b) The inverse direction for generating new samples.}
\label{coupling}
\vspace{-10pt}
\end{figure}

The composed bijective transformations are $L$ conditional affine coupling layers. As shown in Figure \ref{coupling}, in the forward transformation, each layer splits the input vector $\vu$ into two factors $[\vu_1, \vu_2]$. Note that in the first coupling layer, the input is $\vx$. Within each coupling layer, the internal functions $\vs_1(\cdot)$, $\vs_2(\cdot)$, and $\vt_1(\cdot)$, $\vt_2(\cdot)$ are formulated as:
\begin{equation}
\begin{gathered}
  \vv_1 = \vu_1 \odot \text{exp}\big(s_1([\vu_2,\va])\big) + t_1([\vu_2,\va]), \\
  \vv_2 = \vu_2 \odot \text{exp}\big(s_2([\vv_1,\va])\big) + t_2([\vv_1,\va]),
\end{gathered}
\end{equation}
where $\odot$ is the element-wise multiplication and the outputs $\vv_1$ and $\vv_2$ are fed into the next affine transformation.  In the last coupling layer, the output will be $\vz$. In the inverse direction, the conditional affine coupling module takes $\vv_1$ and $\vv_2$ as inputs:
\begin{equation}
    \begin{split}
    &\vu_2 = \big(\vv_2 - t_2([\vv_1,\va])\big) \oslash \text{exp}\big(s_2([\vv_1,\va])\big), \\
    &\vu_1 = \big(\vv_1 - t_1([\vu_2,\va])\big) \oslash \text{exp}\big(s_1([\vu_2,\va])\big),
    \end{split}
\end{equation}
where $\oslash$ is the element-wise division. \noindent As proposed in Real NVP \cite{dinh2016density}, when combining coupling layers, the Jacobian determinant remains tractable and its logarithm is the sum of $s_1(\cdot)$ and $s_2(\cdot)$ over visual feature dimensions.

\noindent \textbf{Discussion: Comparison to IZF.} 
Comparing to state-of-the-art approach IZF \cite{shen2020invertible} that also leverages a conditional generative flow in GZSL, our proposed GSMFlow differs in two aspects: 

1) \textit{Explicit conditional encoding.} In the training stage, IZF learns conditional encoding by disentangling the visual features into semantic and non-semantic vectors:
\begin{equation}
\begin{gathered}
  \text{[}\vz,\va \text{]} = f(\vx, \theta).
\end{gathered}
\end{equation} 
We argue that this approach of incorporating conditional information is implicit and may cause the generated samples to be incoherent with the given attributes and deviated from the real distribution. Instead, we explicitly blend the semantic information into each of the bijective coupling layers. By repeatedly enhancing the impact of the semantic information in the model, the generated visual samples tend to be more semantically consistent.

2) \textit{Drawbacks of the negative MMD.} In IZF, a negative Maximum Mean Discrepancy (MMD) regularization is applied to increase the discrepancy between the generated unseen distributions and the real distributions of the seen classes. We argue that this regularization is infeasible for the two mixed distributions of seen and unseen classes. MMD is commonly used in the domain adaptation that minimizes the discrepancy between the source and the target domains. However, in GZSL, as the seen and unseen classes generally come from the same domain, their visual relationships are highly coalesced. Hence, generally separating the two distributions may cause unexpected distribution shifts, which is undesirable for generating realistic unseen samples. 

In addition to the above differences, we also identify two problems, \textit{i.e.,} variance decay and structural permutation. Accordingly, the visual perturbation and the relation positioning strategies are proposed to handle the two problems.

\subsection{Visual Perturbation}
To mitigate the variance decay problem, we introduce a simple yet effective visual perturbation strategy to preserve the intra-class variance. We begin with sampling a perturbation vector $\ve \in \mathbb{R}^{d_v}$ from Gaussian distribution with the same size $d_v$ as the visual features:
\begin{equation}
\begin{gathered}
  \ve \sim \mathcal{N}(\bm{0},\bm{I}).
\end{gathered}
\end{equation}
In the high-dimensional visual space, we aim to make the perturbation more diverse. Hence, we construct a dropout mask $\vm$ to selectively filter out some of the dimensions of the perturbation vectors:
\begin{equation}
\begin{gathered}
  \vm_j = \begin{cases}
\bm{1}, \quad \text{with}~ p_{drop}
\\
\bm{0}, \quad \text{otherwise}
\end{cases}
\end{gathered}
\end{equation}
where $p_{drop}$ is the probability of keeping the perturbation in the $j$-th dimension. Then, we have the perturbation vector after applying the dropout map:
\begin{equation}
\begin{gathered}
\hat{\ve} = \ve \cdot \vm.
\end{gathered}
\end{equation}
\noindent A virtual sample $\vx^{vir}$ can be then yielded by:
\begin{equation}
\begin{gathered}
  \vx^{vir} = \vx + \lambda_1 \hat{\ve},
\end{gathered}
\end{equation}
where $\lambda_1$ is the coefficient that controls the degree of perturbation.

While the visual perturbation reveals the underlying virtual samples, we may unexpectedly incorporate some noisy samples resulting in the distribution shifts. We argue that the prototype of each class should be invariant when introducing virtual samples. Thus, in order to avoid distribution shift, we aim to fix the prototype for each class when perturbing the real samples. A class prototype is defined as the mean vector of the samples in the same class from the empirical distribution $p_{\bar{X}}$:
\begin{equation}
\begin{gathered}
  \mathbb{E}_{\vx_c \sim p_{\bar{X}}^{c}}[\bar{\vx}_c] = \frac{1}{N_c} \sum_{i=1}^{N_c}\vx_c^i,
\end{gathered}
\end{equation}
where $N_c$ is the sample number of the $c$-th class in the training set.

When generating a visual sample conditioned on $c$-th class with the corresponding semantic vector $\va_c$, the expected mean sample should be close to the class prototype given the prior $\vz$ as the mean vector from the prior distribution, \textit{i.e.,} all zeros $\bm{0}$:
\begin{equation}
\begin{gathered}
  \mathcal{L}_{proto} = \frac{1}{C_{s}} \sum^{C_{s}}_{c=1} \lVert f^{-1}(\bm{0},\va_{c}) - \mathbb{E}_{\vx_c \sim p_{\bar{X}}^{c}}[\bar{\vx}_c] \rVert^{2},
\label{proto}
\end{gathered}
\end{equation}
where $C_{s}$ is the number of seen classes. 

\subsection{Relative Positioning}
\label{gs}
To enhance the inter-class relationship for unseen classes, we introduce the relative positioning technique, preserving the geometric information between different attribute categories in the shared subspace. 
We revise the semantic vectors by measuring the responses to a particular number of semantic anchors. The revised semantic vectors are defined as the global semantic vectors.
We begin with constructing a semantic graph with the class-level semantic vectors. The edges $\mathcal{E}$ are defined as the cosine similarities between all semantic vectors:
\begin{equation}
\begin{gathered}
  \mathcal{E}_{ij} = \frac{\va_i\cdot\va_j}{\lVert\va_i \rVert_2 \cdot \lVert\va_j \rVert_2},
\end{gathered}
\end{equation}
where $\mathcal{E}_{ij}$ refers to the similarity between $i$-th class and $j$-th class. Then, for each class, we calculate the sum similarities to all other classes:
\begin{equation}
\begin{gathered}
  \vd_{i} = \sum_{j=1}^{C_s}\mathcal{E}_{ij},
\end{gathered}
\end{equation}
where $C_s$ denotes the total number of seen classes. We define the three semantic anchors $\va_{max}$,  $\va_{min}$, $\va_{med}$ with the highest, lowest, and median sum similarities to other semantic vectors. The global semantic vectors are then acquired by computing the responses from these three semantic anchors.

The dimensionality of visual features is usually much higher than semantic vectors, \textit{e.g.,} 2,048 vs. 85 in the AWA dataset. Thus, the generation process can be potentially dominated by visual features. The Section \ref{hpa} discusses the impact of the semantic vectors' dimensionality. To avoid this issue, we apply three functions $h_{max}(\cdot),h_{min}(\cdot),h_{med}(\cdot)$ to map the semantic responses to a higher dimension. The global semantic vector for each class can then be formulated as:
\begin{equation}
\begin{gathered}
  \va^{g}_{i} =  h_{max}(\va_{i} - \va_{max}) + h_{min}(\va_{i} - \va_{min}) + h_{med}(\va_{i} - \va_{med}).
\end{gathered}
\end{equation}
The global semantic vectors revised by relative positioning are fed into the conditional generation flow as the conditional information.

\subsection{Training and Zero-shot Inference}

With the derived virtual samples $\vx^{vir}$ and the global semantic vectors $\va^{g}$, the objective functions $\mathcal{L}_{flow}$ and $\mathcal{L}_{proto}$ in Equation \ref{flow} and Equation \ref{proto} should be rewritten as:
\begin{equation}
\begin{gathered}
\mathcal{L}_{flow} = \mathbb{E}[-\log p_{X}(\vx^{vir};\va^{g},\theta)] - \log(p_{\theta}(\theta)), \\
\mathcal{L}_{proto} = \frac{1}{C_{s}} \sum^{C_{s}}_{c=1} \lVert f^{-1}(\bm{0},\va^{g}_{c}) - \mathbb{E}_{\vx_c \sim p_{\bar{X}}^{c}}[\bar{\vx}_c] \rVert^{2}.
 \end{gathered}
\end{equation}

Then, the overall objective function of the proposed GSMFlow is formulated as:
\begin{equation}
\begin{gathered}
  \underset{\theta}{min}( \mathcal{L}_{flow} + \lambda_2 \mathcal{L}_{proto}),
\end{gathered}
\end{equation}
where $\lambda_2$ is the coefficient of the prototype loss.

After the conditional generative flow is trained on the seen classes with the virtual samples and the global semantic vectors, it is leveraged to generate visual features of unseen classes:
\begin{equation}
\begin{gathered}
\bar{\vx}^{u} = g(\vz;\va^g, \theta).
\end{gathered}
\end{equation}
A softmax classifier is then trained on the real visual features of seen classes and the synthesized visual features of unseen classes. For the coming test samples from either seen or unseen classes, the softmax classifier aims to predict the corresponding class label accurately.

\section{Experiments}

\begin {table*}[t]
\caption {Performance comparison in accuracy (\%) on four datasets. We report the accuracies of unseen, seen classes and their harmonic mean, which are denoted as U, S and H. The best results of the harmonic mean are highlighted in bold. $\dag$ and $\ddag$ represent embedding-based and generative methods, respectively.}
\centering
\scalebox{1.15}{
\begin{tabular}[t]{c|c|ccc|ccc|ccc|ccc}
\specialrule{.1em}{.00em}{.00em}
 \rowcolor[gray]{.9} \multirow{2}{*}{} &  \multirow{2}{*}{}   & \multicolumn{3}{c|}{aPaY} & \multicolumn{3}{c|}{AWA}  &  \multicolumn{3}{c|}{CUB} & \multicolumn{3}{c}{FLO}  \\  
  
 \rowcolor[gray]{.9} &  Methods & \textit{U} & \textit{S} & \textit{H}  & \textit{U} & \textit{S} & \textit{H}  & \textit{U} & \textit{S} & \textit{H} & \textit{U} & \textit{S} & \textit{H}\\

  \specialrule{.1em}{.00em}{.00em}
\multirow{6}{*}{\dag}   
                      &   LATEM   \cite{xian2016latent}  
    & 0.1           & 73.0              & 0.2        
    & 13.3          & 77.3              & 20.0
    & 15.2          & 57.3              & 24.0    
    & 6.6           & 47.6              & 11.5  
    \\
 
                      &   ALE    \cite{akata2015label}   
    & 4.6           & 73.7              & 8.7      
    & 14.0          & 81.8              & 23.9    
    & 23.7          & 62.8              & 34.4    
    & 13.3          & 61.6              & 21.9  
    
    \\
                   &   SJE       \cite{akata2015evaluation}
    & 1.3          & 71.4              & 2.6       
    & 8.0          & 73.9              & 14.4  
    & 23.5         & 59.2              & 33.6      
    & -            & -                 & -    
    \\

                    & SAE     \cite{kodirov2017semantic}  
    & 0.4           & 80.9              & 0.9        
    & 1.1           & 82.2              & 2.2 
    & 7.8           & 54.0              & 13.6
    & -             & -                 & -    
    \\
                    & LFGAA       \cite{liu2019attribute}   
    & -             & -                 & -       
    & 27.0          & 93.4              & 41.9  
    & 36.2          & 80.9              & 50.0      
    & -             & -                 & -    
    \\

                    & TCN       \cite{jiang2019transferable}
    & 24.1          & 64.0              & 35.1       
    & 61.2          & 65.8              & 63.4
    & 52.6          & 52.0              & 52.3      
    & -             & -                 & -    
    \\
                    & DVBE      \cite{min2020domain}
    & 32.6          & 58.3              & 41.8          
    & 63.6          & 70.8              & 67.0     
    & 53.2          & 60.2              & 56.5         
    & -             & -                 & -   
    \\

 \specialrule{.1em}{.20em}{.20em}
\multirow{10}{*}{\ddag} & GAZSL \cite{zhu2018generative}      
    & 14.2          & 78.6              & 24.0      
    & 35.4	        & 86.9              & 50.3   
    & 31.7	        & 61.3	            & 41.8       
    & 28.1          & 77.4              & 41.2 
    \\

                        & f-CLSWGAN \cite{xian2018feature}     
   & 32.9           & 61.7              & 42.9    
   & 56.1           & 65.5              & 60.4     
   & 43.7           & 57.7              & 49.7	        
   & 59.0           & 73.8              & 65.6 
   \\
                     &  CANZSL   \cite{chen2020canzsl}         
    & -                 & -                 & -  
    & 49.7              & 70.2              & 58.2  
    & 47.9              & 58.1              & 52.5 
    & 58.2              & 77.6              & 66.5 
    \\
 
                    & CADA-VAE  \cite{schonfeld2019generalized}
    & 31.7              & 55.1              & 40.3          
    & 55.8              & 75.0              & 63.9     
    & 51.6              & 53.5              & 52.4     
    & 51.6              & 75.6              & 61.3 
    \\        
                    & f-VAEGAN-D2  \cite{xian2019f} 
    & -                 & -                 & -         
    & 57.6              & 70.6              & 63.5     
    & 48.4              & 60.1              & 53.6     
    & 56.8              & 74.9              & 64.6 
    \\       
                    & EUC-VAE  \cite{chen2021entropy}
    & 35.0              & 62.7              & 44.9     
    & 55.2              & 78.9              & 64.9  
    & 50.8              & 55.1              & 52.9       
    & 54.0              & 79.0              & 64.1  
    \\
             
                    & TF-VAEGAN  \cite{narayan2020latent}
    & -                 & -                 & -     
    & 59.8              & 75.1              &  66.6  
    & 52.8              & 64.7              & 58.1       
    & 62.5              & 84.1              & 71.7  
    \\

                    & E-PGN  \cite{yu2020episode}
    & -                 & -                 & -          
    & 52.6              & 83.5              & 64.6    
    & 52.0              & 61.1              & 56.2         
    & 71.5              & 82.2              & 76.5  
     
    \\ 
        & IZF \cite{shen2020invertible}
    & 42.3              & 60.5              & 49.8          
    & 60.6              & 77.5              & 68.0    
    & 52.7              & 68.0              & 59.4    
    & -              & -              & -  
     
    \\

\specialrule{.1em}{.00em}{.00em}

\rowcolor[gray]{.9} &   GSMFlow
& 42.0 & 62.3 & \textbf{50.2}  
& 64.5 & 82.1 & \textbf{72.3}
& 61.4 & 67.4 & \textbf{64.3}
& 86.6 & 87.8 & \textbf{87.2} 
\\
\specialrule{.1em}{.00em}{.00em}
\end{tabular}}
\label{gzslperoformance}
\end {table*} 

In this section, we evaluate our approach GSMFlow in both generalized zero-shot learning and conventional zero-shot learning tasks. We first introduce the datasets and experimental settings and then compare GSMFlow with the state-of-the-art methods. Finally, we study the effectiveness of the proposed model with a series of ablation study and hyper-parameter sensitivity analysis. 

\subsection{Datasets}
We conduct experiments on four widely used benchmark datasets of image classification. They are two fine-grained datasets, \textit{i.e.,} Caltech-UCSD Birds-200-2011 (\textbf{CUB}) \cite{wah2011caltech} and Oxford  Flowers (\textbf{FLO}) \cite{nilsback2008automated}, two coarse-grained datasets, \textit{i.e.,} Attribute Pascal and Yahoo (\textbf{aPaY}) \cite{farhadi2009describing} and Animals with Attributes 2 (\textbf{AWA}) \cite{lampert2013attribute}.
CUB consists of 11,788 images from 200 fine-grained bird species, in which 150 selected as seen classes and 50 as unseen classes. For FLO, it contains 8,189 images from 102 flower categories, 82 of which are chosen as seen classes. The class-level semantic vectors of CUB and FLO are extracted from the fine-grained visual descriptions (10 sentences per image), yielding 1,024-dimensional character-based CNN-RNN features \cite{reed2016learning} for each class. aPaY dataset contains 18,627 images from 42 classes. There are 30 seen classes and 12 unseen classes respectively. Each class is annotated with 64 attributes. AWA2 is a considerably larger dataset with 30,475 images from 50 classes and they are annotated with 85 attributes. For the split, 40 of the total classes are selected as seen classes and ten as unseen classes. For each dataset, We follow the proposed split setting in \cite{xian2018zero} and \cite{nilsback2008automated}. 

\subsection{Implementation Details}
Our framework is implemented with the open-source machine learning library PyTorch. The conditional generative flow consists of a series of affine coupling layers. Each affine coupling layer is implemented with two fully connected (FC) layers and the first FC layer is followed by a LeakyReLU activation function. The hidden dimension of the FC layer is set as 2,048. The coefficients $\lambda_1$ of perturbation degree and the coefficient $\lambda_2$ of the prototype loss are set within \{0.02, 0.05, 0.15, 0.3, 0.5\} and \{1, 3, 10, 20, 30\}. The dimension of global semantic vectors varies in \{128, 256, 512, 1,024, 2,048\} and the number of the affine coupling layers varies in \{1, 3, 5, 10, 20\}.  The corresponding results are given in Section \ref{hpa}. The function $h(\cdot)$ is implemented with an FC layer and a ReLU activation function. We use Adam optimizer with $\beta_{1}=0.9$, $\beta_{2}=0.999$ and set the batch size to 256 and set the learning rate to 3e-4. All the experiments are performed on a Lenovo workstation with two NVIDIA GeForce GTX 2080 Ti GPUs.

\subsection{Evaluation Metrics}
To avoid the failure of classification accuracy for imbalanced class distributions, we adopt average per-class Top-1 accuracy as the fair evaluation criteria for conventional ZSL and the seen and unseen set performance in GZSL:
\begin{equation}
\begin{aligned}
  Acc_{\mathcal{Y}} = \frac{1}{|\mathcal{Y}|} \sum^{|\mathcal{Y}|}_{y=1} \frac{\# \  of \ correct \ predictions \ in \ y}{\# \ of \ samples \ in \ y},
\end{aligned}
\end{equation}
where $|\mathcal{Y}|$ is the number of testing classes. A correct prediction is defined as the highest probability of all candidate classes.
Following \cite{xian2018zero}, the harmonic mean of the average per-class Top-1 accuracies on seen $Acc_S$ and unseen $Acc_U$ classes are used to evaluate the performance of generalized zero-shot learning. It is computed by:
\begin{equation}
\begin{aligned}
  H = \frac{2*Acc_S*Acc_U}{Acc_S+Acc_U}.
\end{aligned}
\end{equation}

\subsection{Comparisons with State-of-the-art Methods}
\label{GZSL}

Table \ref{gzslperoformance} summarizes the performance comparison between our proposed GSMFlow and other state-of-the-art methods in the setting of GZSL. We choose the most representative state-of-the-art methods for comparison, the top eight methods marked with $\dag$ are embedding-based methods, and the below eight methods and our proposed method marked with $\ddag$ are generative methods. It can be seen that our proposed framework consistently outperforms other methods.
Among the generative methods, IZF \cite{shen2020invertible} also leverages normalizing flows as the base generative model. 
However, we show that by mitigating the generation shifts, our proposed GSMFlow achieves significant improvement on these four datasets.


\begin {table}[t]
\caption {Conventional ZSL accuracy (\%). The best results are formatted in bold.}
\begin{center}
\scalebox{1.}{
\begin{tabular}[h]{ l | c | c | c | c  }
\specialrule{.1em}{.00em}{.00em}
   \rowcolor[gray]{.9}  Method & aPaY & AWA & CUB & FLO \\
  \specialrule{.1em}{.00em}{.00em}

  ALE   \cite{akata2015label}             
                    & 39.7      & 62.5      & 54.9          & 48.5          \\
  SJE   \cite{akata2015evaluation}          
                    & 31.7      & 61.9      & 54.0          & 53.4          \\
  ESZSL \cite{romera2015embarrassingly}           
                    & 38.3      & 58.6      & 51.9          & 51.0          \\
  LFGAA \cite{liu2019attribute}   
                    & -         & 68.1      & 67.6          & -          \\
  DCN \cite{liu2018generalized}
                    & 43.6      & 65.2      & 56.2          & -          \\
  TCN \cite{jiang2019transferable}
                    & 43.6      & 65.2      & 56.2          & -          \\
  GAZSL \cite{zhu2018generative}      
                    & 41.1      & 70.2      & 55.8          & 60.5	        \\
  f-CLSWGAN \cite{xian2018feature}     
                    & 40.5      & 65.3      & 57.3          & 69.6          \\
  cycle-CLSWGAN  \cite{felix2018multi}         
                    & -         & 66.8      & 58.6          & 70.3          \\
  CADA-VAE  \cite{schonfeld2019generalized}
                    & -         & 64.0      & 60.4          & -             \\        
  f-VAEGAN-D2  \cite{xian2019f} 
                    & -         & 71.1      & 61.0          & 67.7          \\       
  DLFZRL  \cite{tong2019hierarchical}
                    & 46.7      & 70.3      & 61.8          & -             \\       
  TF-VAEGAN  \cite{narayan2020latent}
                    & -         & 72.2      & 64.9          & 70.8           \\
  E-PGN  \cite{yu2020episode}
                    & -         & 73.4      & 72.4          & 85.7           \\
  IZF  \cite{shen2020invertible}
                    & 44.9         & \textbf{74.5}      & 67.1          & -           \\
\specialrule{.1em}{.00em}{.00em}
 \rowcolor[gray]{.9} GSMFlow          & \textbf{49.2}    & 72.7      & \textbf{76.4}          & \textbf{86.9}  \\
\specialrule{.1em}{.00em}{.20em}
\end{tabular}}
\end{center}
\label{czsl}
\end {table}

\subsection{Conventional Zero-shot Learning}
\label{ZSL}
Even if GSMFlow is mainly proposed for GZSL, to further validate the effectiveness of our proposed framework, we also conduct experiments in the context of conventional ZSL. Table \ref{czsl} summarizes the conventional zero-shot learning performance. We achieve better performance than all the compared methods on the aPaY, CUB, and FLO datasets. For the AWA dataset, we can also achieve comparatively good performance. 

\begin {table}[t]
\caption {Effects of different components on aPaY and AWA datasets. U, S and H represent unseen, seen and harmonic mean, respectively. The best results of harmonic mean are formatted in bold.}
\begin{center}
\scalebox{0.91}{
\begin{tabular}[t]{c|ccc|ccc}
\specialrule{.1em}{.00em}{.00em}
  \rowcolor[gray]{.9}     & \multicolumn{3}{c|}{aPaY} & \multicolumn{3}{c}{AWA}   \\  \rowcolor[gray]{.9} Settings
   & \textit{U} & \textit{S} & \textit{H}  & \textit{U} & \textit{S} & \textit{H}  \\

    \specialrule{.1em}{-.15em}{.00em}
  IZF w/o constraints      & 35.2      & 54.2     & 42.7      
            & 38.1      & 78.9      & 51.4   
            \\

  GSMFlow w/o constraints   
            & 36.5      & 60.0      & 45.4       
            & 53.3      & 67.6      & 59.6   
            \\

  GSMFlow w/o VP     
            & 38.4      & 62.3      & 47.6        
            & 62.2      & 71.3      & 66.4   
            \\

 GSMFlow w/o RP
            & 39.6      & 61.3      & 48.1         
            & 62.9      & 80.4      & 70.6
            \\

    \specialrule{.1em}{.10em}{.00em}

 \rowcolor[gray]{.9} GSMFlow      
& 42.0 & 62.3 & \textbf{50.2}  
& 64.5 & 82.1 & \textbf{72.3}
\\

\specialrule{.1em}{.00em}{.00em}
\end{tabular}}
\end{center}
\label{ablation}
\end {table}

\begin{figure}[t]
\centering
\includegraphics[width=85mm]{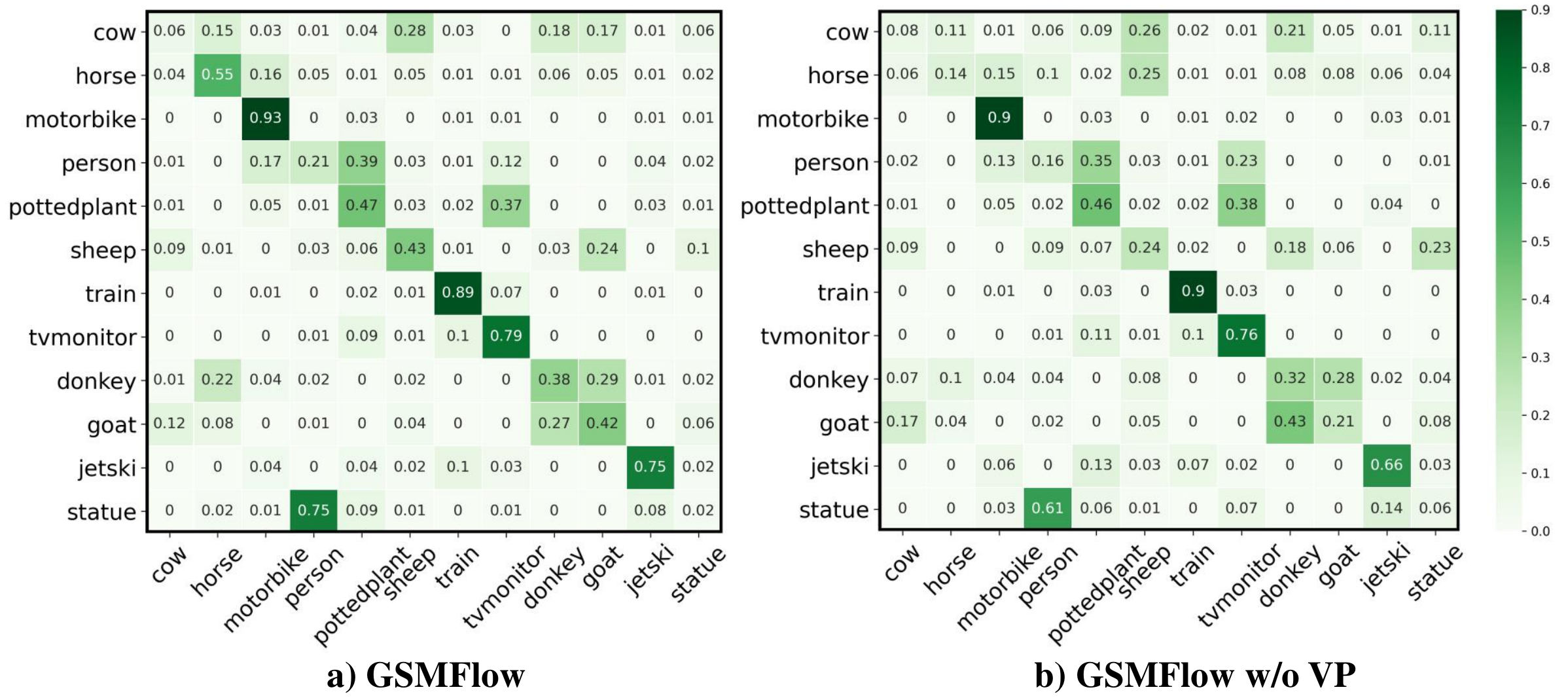}
\caption{Class-wise performance comparison between with and without visual perturbations. The vertical labels are the groundtruth and the horizontal labels are the predictions.}
\label{cm}
\end{figure}

\subsection{Ablation Study}
To analyze the contribution of each proposed component and the merit from addressing the three problems in generative zero-shot learning, \textit{i.e., semantic inconsistency, variance dacay}, and \textit{structural permutation}, we conduct an ablation study on the proposed GSMFlow. We decompose the complete framework into four variants. These include: \textit{\textbf{IZF w/o constraints}} - the conditional generative flow adopted in compared method \cite{shen2020invertible}; \textit{\textbf{GSMFlow w/o constraints}} - the generative flow in our proposed framework without visual perturbation and global semantic learning; \textit{\textbf{GSMFlow w/o VP}} - the complete framework without visual perturbation; and \textit{\textbf{GSMFlow w/o RP}} - the original class-level semantic vectors are used. 

Comparing between \textit{\textbf{IZF w/o constraints}} and \textit{\textbf{GSMFlow w/o constraints}}, we can see the performance comparison between the two ways of incorporating conditional information in the generative flow. It can be seen that, instead of mixing the conditional information with the prior input in IZF, our explicit conditional strategy that progressively injects the semantic information into the affine coupling layers during training can achieve higher accuracy on GZSL. 
The results on the variants \textit{\textbf{GSMFlow w/o VP}} shows that by mitigating the structural permutation issue, the relative positioning strategy that captures geometric relationships between semantic vectors can significantly improve the GZSL performance.
\textit{\textbf{GSMFlow w/o RP}} indicates that effectiveness of the visual perturbation strategy to prevent the variance decay. When combining these components together, we achieve the best performance results.

\begin{figure*}[t]
\centering
\includegraphics[width=180mm]{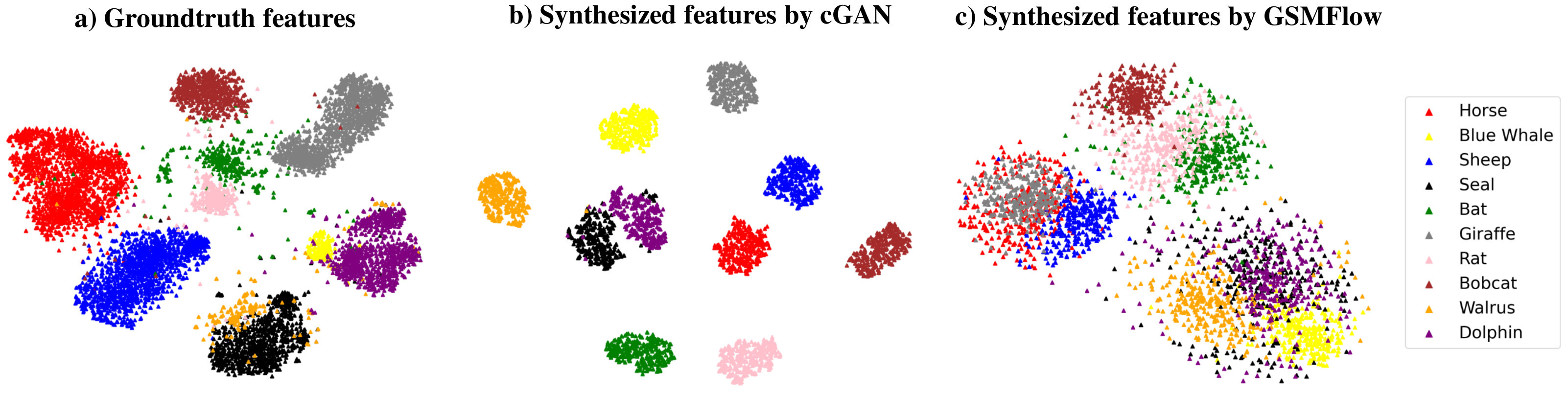}
\caption{Distribution comparison of the unseen classes in the AWA dataset. (a) The real distributions of the visual features extracted from the backbone. (b) The synthesized distributions by cGAN. (c) The synthesized distributions by GSMFlow.}
\label{tsne}
\end{figure*}

\begin{figure}[t]
\centering
\includegraphics[width=85mm]{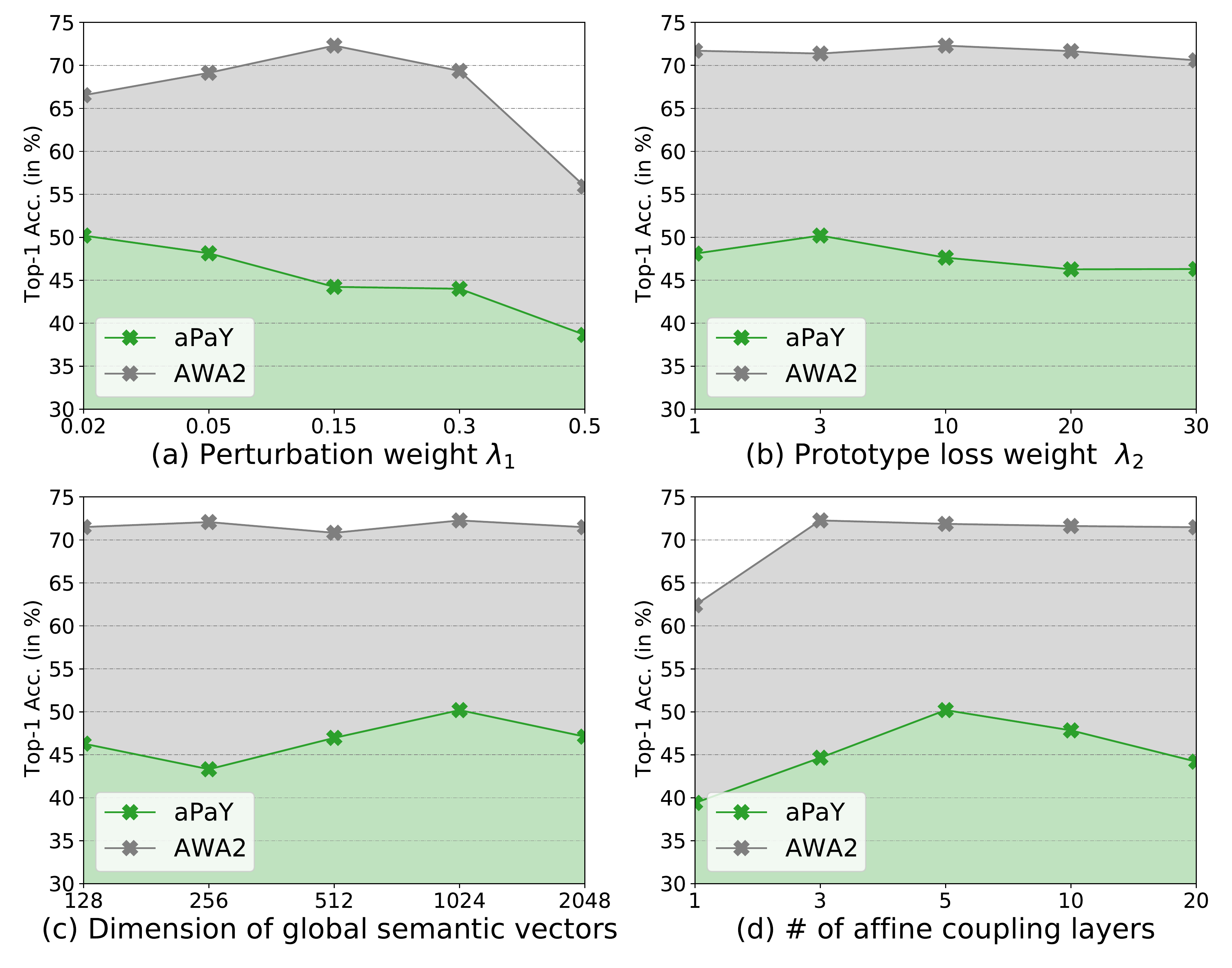}
\caption{Hyper-parameter sensitivity. The horizontal axis indicates the varying hyper-parameters for (a) perturbation weight, (b) prototype loss weight, (c) semantic vector dimensions, and (d) number of affine coupling layers. The vertical axis reports the corresponding performance.}
\label{hyper}
\end{figure}

To further investigate the performance boost through visual perturbation, in Figure \ref{cm}, we compare the class-wise accuracy with and without visual perturbation by illustrating the confusion matrices in the two settings. Without visual perturbation, we can notice that \textit{horses} can be easily misclassified as \textit{sheep} and can only achieve 14\% accuracy. The performance surges to 55\% when we introduce the visual perturbation strategy. Similar observations hold for other classes.

\subsection{Hyper-parameter Analysis}
\label{hpa}
GSMFlow mainly involves four hyper-parameters in the model training, as shown in Figure \ref{hyper}. Varying the perturbation weight $\lambda_1$ from 0.02 to 0.5, in Figure \ref{hyper}(a), we find that the aPaY and AWA can reach the peak at around 0.02 and 0.15, respectively. 
It turns out that the weight of the prototype loss $\lambda_2$ does not have a significant impact on the performance.
In Figure \ref{hyper}(b), we vary $\lambda_2$ between 1 and 30, and observe that at 3 and 10, the best performance results are achieved on aPaY and AWA2. 
As discussed in Section \ref{gs}, the semantic vectors usually have lower dimensions than the visual features, which makes the visual features dominate the generation process in the affine coupling layers. 
In Figure \ref{hyper}(c), we report the impact from the dimensionality of the semantic vectors. It can be seen that low-dimensional semantic vectors tend to jeopardize the performance. The best performance results are both achieved at 1,024 dimensions. 
We also investigate the impact on the number of conditional affine coupling layers, which directly influences the generative model size. In Figure \ref{hyper}(d), we can see the best performance results are from 3 and 5 layers respectively for the aPaY and AWA datasets.

\subsection{t-SNE visualization}
The quantitative results reported for GZSL (Section \ref{GZSL}) and ZSL (Section \ref{ZSL}) demonstrate that the visual samples of the unseen classes generated by GSMFlow are of good quality and effective for classification tasks. 
To further gain an insight into the quality of the generated samples and validate the motivation illustrated in Figure \ref{intro}, we compare the empirical distribution of the real unseen class data and the synthesized unseen class data by a conditional generative adversarial network (cGAN) and our proposed GSMFlow, as depicted in Figure \ref{tsne}. 
It can be seen that the distributions of the synthesized unseen data from cGAN suffer from the generation shifts problem, which is undesirable for approximating optimal decision boundaries. Specifically, each class is collapsed to fixed modes without capturing enough variances. 
In contrast, thanks to the visual perturbation strategy, the synthesized samples by GSMFlow are much more diverse, which helps to approximate optimal decision boundaries. The class-wise relationship is also reflected. 
For example, comparing to other animal species, \textit{horse} and \textit{giraffe} share some attribute values. Also, all the marine species, \textit{blue whale, walrus, dolphin, and seal} are well separated from other animals. 
As a result, the generated samples are semantically consistent.

\section{Conclusion}
In this paper, we propose a novel Generation Shifts Mitigating Flow (GSMFlow) framework, which is comprised of multiple conditional affine coupling layers for learning unseen data synthesis. There are three potential problems that trigger the generation shifts for this task, \textit{i.e., semantic inconsistency, variance decay}, and the \textit{structural permutation}. First, we explicitly blend the semantic information into the transformations in each of the coupling layers, reinforcing the correlations between the generated samples and the corresponding attributes. Second, a visual perturbation strategy is introduced to diversify the generated data and hereby help adjust the decision boundary of classifier. Third, to avoid structural permutation in the semantic space, we propose a relative positioning strategy to manipulate the attribute embeddings, guiding which to fully preserve the inter-class geometric structure. An extensive suite of experiments and analysis show that MGSFlow can outperform existing generative approach for GZSL that suffers from the problems of the generation shifts.

\newpage
\balance{
\bibliographystyle{ACM-Reference-Format}
\bibliography{files/bib/bibliography.bib}}

\end{document}